%% file: main.tex
\pdfoutput=1
\documentclass[letterpaper, 10 pt, conference]{ieeeconf}

\IEEEoverridecommandlockouts
\usepackage{cite}

\usepackage{graphicx}
\usepackage{amsmath,amssymb,amsfonts,bm}
\usepackage{mathtools}
\usepackage[protrusion=false]{microtype}
\usepackage{booktabs}
\usepackage{multirow}
\usepackage{array}
\usepackage{tabularx}
\usepackage{xcolor}
\usepackage{url}
\usepackage{xspace}
\let\labelindent\relax
\usepackage{enumitem}

\usepackage{algorithm}
\usepackage{algorithmic}
\usepackage{float}
\usepackage{placeins}

\newcommand{\IEEEtabtoppad}{\vspace*{8pt}}
\makeatletter
\newcommand{\IEEEalgtoppad}{%
  \gdef\@fs@pre{\vbox{\kern8pt}\hrule height.8pt depth0pt \kern2pt}}
\makeatother

\usepackage[hidelinks]{hyperref}
\usepackage[nameinlink,noabbrev]{cleveref}

\graphicspath{{Fig/}{fig/}{./}}
\renewcommand{\keywords}[1]{}

\newcommand{\calD}{{\cal D}}

\newcommand{\calH}{{\cal H}}

\newcommand{\calM}{{\cal M}}

\newtheorem{definition}{Definition}
\newtheorem{remark}{Remark}
\newtheorem{example}{Example}
\newtheorem{problem}{Problem}

\title{\LARGE \bf Dual Gradient Consensus for Adaptive Learning to Transient Out-of-Distribution Disturbances in Quadrotor Control}

\author{Mingtian Tan and Vasileios Tzoumas%
\thanks{This work has been submitted to the IEEE for possible publication. Copyright may be transferred without notice, after which this version may no longer be accessible.}%
}

\begin{document}

\makeatletter
\def\@textbottom{\vskip \z@ \@plus 0.4\baselineskip}
\makeatother

\maketitle
\thispagestyle{empty}
\pagestyle{empty}

\input{sec/0-Abstract}
\input{sec/1-Introduction}
\input{sec/2-Problem}
\input{sec/3-Algo}
\input{sec/4-Exp}
\input{sec/5-Conclusion}

\input{sec/6-Appendix}

\bibliographystyle{IEEEtran}
\vspace{-4mm}
\bibliography{reference}

\end{document}

%% file: sec/0-Abstract.tex
\begin{abstract}
Quadrotor trajectory tracking under unmodeled disturbances often relies on online residual adaptation, but short out-domain residuals can corrupt the learned disturbance model when constant-step updates treat every residual gradient uniformly.
We propose Dual Gradient Consensus~(DGC), an adaptive-learning method for mixed in-domain/out-domain disturbances that retains the random Fourier feature model, offline BPTT policy, and least-squares adaptation of a prior domain-adaptive residual policy~(DAPL) while modifying the online update step size.
DGC compares each residual gradient with long- and short-term gradient memories, classifies the current adaptation regime, and uses a confidence map to suppress unreliable updates while recovering adaptation after persistent domain shifts.
We evaluate DGC in numerical quadrotor trajectory tracking under four residuals that mix an in-domain component with short out-domain events.
Compared with $L_1$-augmented MPC, RFF-based MPC, constant-step DAPL, and a frozen-parameter baseline, DGC achieves the lowest mean position RMSE in all four cases.
\end{abstract}

%% file: sec/1-Introduction.tex
\section{Introduction}
\label{sec:intro}
Future aerial robots are expected to track planned trajectories accurately in complex and cluttered near-ground settings, including infrastructure inspection, last-mile delivery, search and rescue, precision agriculture, mapping and surveying, and urban air mobility, under unmodeled and often systematically unknown disturbances~\cite{marshall2021survey,cohen2021urban,ahmed2025uam,tacconi2026taxonomy}.
In such settings, high-speed flight brings unexpected transient events---an inadvertent low-energy collision, a gust strike, brief contact with surrounding structures, leaves ingested by a rotor, or an in-flight motor fault---after which the vehicle must still be controllable rather than only regulated about the original plan~\cite{loquercio2021learning,sun2021incremental,mao2024propeller}.

It is challenging to track the planned trajectory accurately while remaining controllable after such transients.
Unknown, time-varying disturbances come in two forms that should not be treated alike: persistent, non-stationary domain changes---changes in aerodynamic loading, a payload that shifts the center of mass in flight, or transitions between regimes with and without ground effect---and the short-lived transients noted above~\cite{tal2020accurate,hanover2021performance,huang2023datt}.
A common remedy is online adaptation of a disturbance estimate or an environment representation, allowing the controller to continuously account for the residual dynamics encountered during deployment~\cite{slotine1991applied,krstic1995nonlinear,ioannou1996robust,wu2023mathcal,kumar2021rma,huang2023datt,zhou2025simultaneous,zhou2026efficient}.
That strategy, however, is itself subject to a mixed-residual issue: the adaptation stream mixes samples that characterize the current in-domain with short out-domain events, and excessive adaptation to the latter causes transient effects to be incorporated into the learned disturbance model rather than treated as temporary deviations.
The remaining issue is therefore how to keep adapting when the domain truly changes, without letting short out-domain residuals rewrite the learned disturbance model.

The current state of the art spans several approaches.
Robust control bounds tracking error against worst-case disturbances without identifying a residual online~\cite{zhou1998essentials,mayne2005robust,zhou2023safe,martin2024guarantees,liu2024robust}.
Classical adaptive control updates coefficients of prescribed basis functions~\cite{slotine1991applied,krstic1995nonlinear,ioannou1996robust}, while disturbance observers, $L_1$ estimators, and related schemes estimate additive residuals for feedback compensation~\cite{tal2020accurate,wu2023mathcal,tao2024robust,das2024robust,hanover2021performance,boffi2021regret,boffi2022nonparametric,jia2023evolver,huang2023datt}, typically through a low-pass frequency split rather than a test of whether the residual is a persistent domain change or a short out-domain residual.
Learning-based controllers train policies over randomized environments~\cite{tobin2017domain,ramos2019bayessim,lee2020learning,li2025reinforcement,sanchez2018real} or condition them on environment encoders, as in RMA and DATT~\cite{kumar2021rma,xue2026robust,zhang2025learning,huang2023datt,farahani2021brief,ben2006analysis}.
Online non-stochastic control updates sequentially under adversarial disturbances~\cite{hazan2022introduction,agarwal2019online,zhao2022non,zhou2023safecdc,zhou2023saferal,zhou2023efficient}.
Predictive residual models have also been used for model-based control~\cite{zhou2025simultaneous,tsiamis2024predictive,zhou2025adaptive}: MPC~(RFF) adapts random-feature coefficients and kernel bandwidth online~\cite{rahimi2007random,rahimi2008uniform,zhou2025simultaneous}, while DAPL applies constant-step least squares to the same pair~\cite{zhou2026efficient}, so in-domain and out-domain residuals enter the update at a uniform rate.
Direct online policy learning on hardware has been shown~\cite{wu2023daydreamer,nan2025efficient,pan2026learning}.
However, residual adaptation typically uses a uniform step.

\begin{figure*}[t]
\centering
\includegraphics[width=\textwidth]{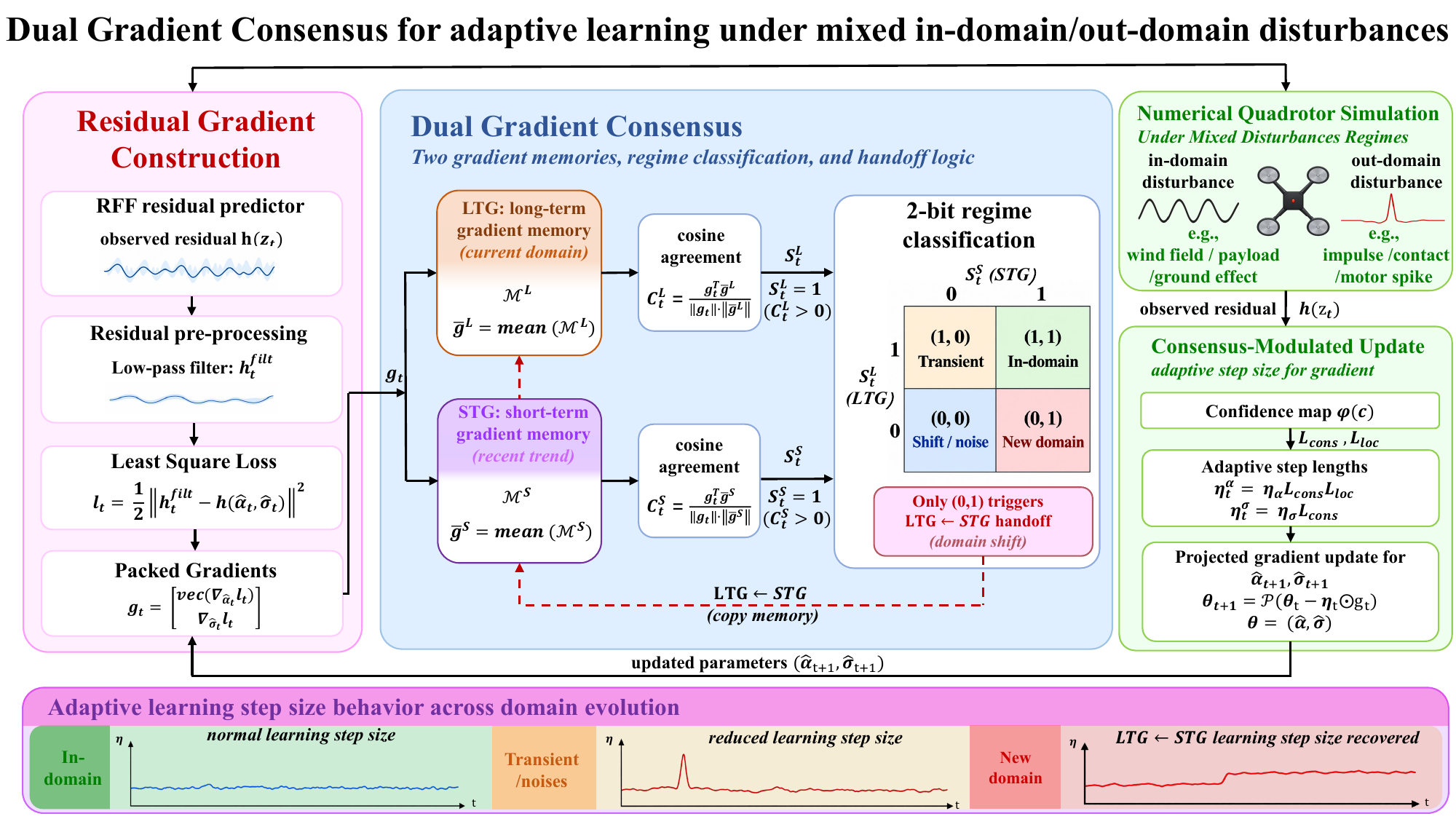}
\caption{Overview of Dual Gradient Consensus for online adaptive learning under mixed in-domain/out-domain disturbances.
Left: the residual pipeline forms a packed gradient $g_t$ from the RFF predictor, a low-pass residual, and the least-squares loss.
Center: Long-Term Gradient (LTG) and Short-Term Gradient (STG) memories classify $g_t$ into four regimes; only $(0,1)$ hands LTG over to STG.
Right: a confidence map sets the step length of the $(\hat{\boldsymbol{\alpha}},\hat{\sigma})$ update on mixed in-domain/out-domain residuals.
Bottom: the applied learning step size remains high in-domain, drops on transients, and recovers after a completed domain shift.}
\label{fig:dgc}
\vspace{-4mm}
\end{figure*}

This paper builds on the main pipeline of Domain-Adaptive Policy Learning~(DAPL)~\cite{zhou2026efficient}---the RFF residual model, offline BPTT policy, and online least-squares loss---and develops a consensus-driven step-size adaptation for updating the RFF parameters that represent the unknown disturbance.
We propose Dual Gradient Consensus~(DGC), overviewed in~\Cref{fig:dgc}.
Two residual-gradient memories, a long-term gradient representing the current domain and a short-term gradient capturing recent evolution, classify each incoming gradient into four adaptation regimes and modulate its update step without altering the gradient itself.
Unlike CAdam, which gates Adam coordinates for neural recommenders when the gradient and momentum disagree~\cite{wang2025cadam}, DGC is a consensus map on the residual step length.
In-domain agreement preserves the long-term memory and applies a consensus-modulated step; disagreement with both memories suppresses the update; persistent agreement with the short-term memory triggers a handoff to the new domain.
Relative to MPC~(RFF)~\cite{zhou2025simultaneous} and DAPL~\cite{zhou2026efficient}, the residual features and least-squares objective remain unchanged, while the kernel parameters continue to adapt online.
The resulting controller retains the same policy as DAPL while limiting the incorporation of short out-domain residuals into the kernel parameters and preserving adaptation to domain shifts.

We evaluate the method in numerical quadrotor trajectory-tracking simulations under mixed in-domain and out-domain disturbances~\Cref{sec:exp}.
Four disturbance profiles combine an in-domain component with impulsive transients: a steady level change, a zero in-domain residual with repeated impulses, a periodic shift, and a linear ramp.
We compare Dual Gradient Consensus with MPC~(L1-E)~\cite{huang2023datt}, MPC~(RFF)~\cite{zhou2025simultaneous}, DAPL~\cite{zhou2026efficient}, and DAPL without $\sigma$ and $\boldsymbol{\alpha}$ updates.
Across all four cases, DGC achieves the lowest mean position RMSE; the advantage is largest under repeated impulses, where constant-step DAPL incorporates transient disturbances into the residual model, while the MPC variants perform well on a stationary offset but degrade under domain shifts.
Time histories of residual prediction error, position tracking error, and the applied step sizes are reported together with the RMSE comparison in~\Cref{subsec:sim}.

%% file: sec/2-Problem.tex
\section{Problem Formulation}
\label{sec:problem}

Consider the quadrotor dynamics as follows,
\begin{equation}
    \dot{\boldsymbol{p}} = \boldsymbol{v}, \;\;  m \dot{\boldsymbol{v}} = m \boldsymbol{g} + \boldsymbol{f} + \boldsymbol{h}, \;\; \dot{\boldsymbol{q}} =\frac{1}{2}\boldsymbol{q}\otimes\left[
        0, \;
        \boldsymbol{\omega}\right]^\top,
    \label{eq:quadrotor}
\end{equation}
where \(\boldsymbol{p}\in\mathbb{R}^{3}\) and \(\boldsymbol{v}\in\mathbb{R}^{3}\) collect inertial-frame position and velocity, \(\boldsymbol{q}\) parametrizes attitude as a quaternion, \(\otimes\) denotes the associated quaternion product, \(\boldsymbol{\omega}\in\mathbb{R}^{3}\) is the angular velocity in the body frame, \(m\) is the mass, \(\mathcal{J}\) is the vehicle inertia matrix, \(\boldsymbol{g}\) is the gravity vector, \(\boldsymbol{f}=\boldsymbol{R}\left[0\ 0\ T\right]^\top\in\mathbb{R}^{3}\) is the rotor thrust written in the inertial frame, \(T\) is the collective thrust along the body \(z\)-axis, and \(\boldsymbol{h}\in\mathbb{R}^{3}\) is the disturbance force. The input to the plant is \([T, \; \boldsymbol{\omega}^\top]^\top\).

We distinguish two classes of disturbances.
The \emph{in-domain} component \(\boldsymbol{d}^{\mathrm{r}}\) captures constant or periodic environment effects that the online learner should adapt to online, such as ground effect, a wind field, or a payload.
The \emph{out-domain} component \(\boldsymbol{d}^{\mathrm{f}}\) captures short-lived transients that should not be written into the online learner, such as a gust strike, a sudden jump or drop induced by motor failure, or other impulsive events.
In the quadrotor residual, these appear as \(\boldsymbol{h}^{\mathrm{reg}}\) and \(\boldsymbol{h}^{\mathrm{off}}\), respectively.

The quadrotor dynamics in \cref{eq:quadrotor} can be compactly written as
\begin{equation}
    \boldsymbol{x}_{t+1} = \boldsymbol{f}(\boldsymbol{x}_{t}) + \boldsymbol{g}(\boldsymbol{x}_{t}) \boldsymbol{u}_{t} + \boldsymbol{h}_{t},
    \label{eq:discrete}
\end{equation}
where \(\boldsymbol{x}\triangleq[\boldsymbol{p}^\top,\boldsymbol{v}^\top,\boldsymbol{q}^\top]^\top\), \(\boldsymbol{u}\triangleq[T,\boldsymbol{\omega}^\top]^\top\), and residual
\begin{equation}
    \boldsymbol{h}_{t}
    \triangleq
    \boldsymbol{x}_{t+1} - \boldsymbol{f}(\boldsymbol{x}_{t}) - \boldsymbol{g}(\boldsymbol{x}_{t})\boldsymbol{u}_{t}
    =
    \boldsymbol{h}^{\mathrm{reg}}_{t} + \boldsymbol{h}^{\mathrm{off}}_{t}
    \label{eq:residual-ood}
\end{equation}
mixes the in-domain residual \(\boldsymbol{h}^{\mathrm{reg}}_{t}\) with the out-domain residual \(\boldsymbol{h}^{\mathrm{off}}_{t}\).
The learner observes only \(\boldsymbol{h}_{t}\); it does not receive a label that separates the two.

\begin{problem}\label{prob:control}
Given a reference trajectory \(\boldsymbol{x}_{\mathrm{ref}}\), the goal is to track it with a control policy while adapting online only an estimate \(\hat{\boldsymbol{h}}^{\mathrm{reg}}\) of the in-domain disturbance, even though the observed residual \(\boldsymbol{h}_t\) mixes an out-domain component \(\boldsymbol{h}^{\mathrm{off}}\), where the disturbance may be time-varying, stationary, or mixed with abrupt out-domain events.
\end{problem}

%% file: sec/3-Algo.tex
\section{Algorithm}
\label{sec:algo}

This section recaps the DAPL pipeline---RFF residual, offline BPTT policy, and online least-squares kernel updates---from~\cite{zhou2026efficient}, then presents Dual Gradient Consensus, which replaces the constant online step size.
\Cref{subsec:algo} states the two online algorithms: DAPL~(\Cref{alg:MODA})~\cite{zhou2026efficient} and Dual Gradient Consensus~(\Cref{alg:DGC}).

\subsection{Function Approximation via Random Fourier Features}\label{subsec:RFF}
We use the same random Fourier feature~(RFF) residual as DAPL~\cite{zhou2026efficient,rahimi2007random,rahimi2008uniform,boffi2022nonparametric}.
Assume \(\boldsymbol{h}:\mathbb{R}^{d_z}\rightarrow\mathbb{R}^{d_x}\) lies in a subspace of an RKHS \(\calH\).
By the operator-valued Bochner theorem,
\(\boldsymbol{h}(\cdot)=\int_\Theta \boldsymbol{\Phi}_\nu(\cdot,\boldsymbol{\theta})\,\boldsymbol{\alpha}(\boldsymbol{\theta})\,\mathrm{d}\nu(\boldsymbol{\theta})\)
and the Monte Carlo truncation
\(\boldsymbol{h}(\cdot)\approx\hat{\boldsymbol{h}}(\cdot;\boldsymbol{\alpha})\triangleq\frac{1}{M}\sum_{i=1}^{M}\boldsymbol{\Phi}_\nu(\cdot,\boldsymbol{\theta}_i)\boldsymbol{\alpha}_i\),
where \(\boldsymbol{\Phi}_\nu(\boldsymbol{z},\boldsymbol{\theta})=\boldsymbol{B}(\boldsymbol{w})\phi(\boldsymbol{w}^{\top}\boldsymbol{z}+b)\),
\(\boldsymbol{B}:\mathbb{R}^{d_z}\rightarrow\mathbb{R}^{d_x\times d_1}\),
\(\phi:\mathbb{R}\rightarrow[-1,1]\) is \(1\)-Lipschitz,
\(d_1\leq d_x\),
\(\boldsymbol{\theta}_i=(\boldsymbol{w},b)\sim\nu\) i.i.d.,
and \(\boldsymbol{\alpha}_i\triangleq\boldsymbol{\alpha}(\boldsymbol{\theta}_i)\) are learned.
As in DAPL, we use a Gaussian kernel of bandwidth \(\sigma\):
\(\boldsymbol{K}(\boldsymbol{x},\boldsymbol{y})=\exp\bigl(-\|\boldsymbol{x}-\boldsymbol{y}\|^2/(2\sigma^2)\bigr)\),
each entry \(\boldsymbol{\Phi}_{ij}\) of \(\boldsymbol{\Phi}_\nu\) equals \(\cos(\sigma\boldsymbol{w}_{ij}^{\top}\boldsymbol{z}+b_{ij})\),
\(\nu=\mathcal{N}(\boldsymbol{0},\sigma^2\boldsymbol{I})\),
each \(\boldsymbol{w}_{ij}\) is standard Gaussian, and each phase \(b\) is uniform on \([0,2\pi]\).

\subsection{Policy Optimization with Analytical Gradients}\label{subsec:BPTT}
As in DAPL~\cite{zhou2026efficient}, the policy is trained offline by Back-Propagation Through Time~(BPTT)~\cite{metz2021gradients} on a differentiable closed-loop rollout.
The \(N\)-step objective is
\begin{equation}
\begin{aligned}
    \max_{\boldsymbol{\psi}} R(\boldsymbol{\psi})
    &=
    \max_{\boldsymbol{\psi}}
    \frac{1}{N}\sum_{i=1}^{N} r(\boldsymbol{x}_i,\boldsymbol{u}_i)\\
    &=
    \max_{\boldsymbol{\psi}}
    \frac{1}{N}\sum_{i=1}^{N} r\bigl(\boldsymbol{x}_i,\pi_{\boldsymbol{\psi}}(\boldsymbol{o}_i)\bigr).
\end{aligned}
\end{equation}
The parameters are then updated with the analytic gradient \(\nabla_{\boldsymbol{\psi}}R\) and step size \(\eta_{\boldsymbol{\psi}}\),
\[
\boldsymbol{\psi}_{k+1}=\boldsymbol{\psi}_{k}+\eta_{\boldsymbol{\psi}}\nabla_{\boldsymbol{\psi}_k}R.
\]

\subsection{Online Least-Squares Estimation}\label{subsec:OLS}
MPC~(RFF) adapts both \(\boldsymbol{\alpha}\) and \(\sigma\) online~\cite{zhou2025simultaneous}; DAPL applies a constant least-squares step to the same pair~\cite{zhou2026efficient}.
Let \(\boldsymbol{z}_t\) denote the RFF input.
Given \(\bigl(\boldsymbol{z}_t,\;\boldsymbol{h}(\boldsymbol{z}_t)\bigr)\), the predictor of \Cref{subsec:RFF} is
$\hat{\boldsymbol{h}}_t\triangleq\frac{1}{M}\sum_{i=1}^{M}\Phi(\boldsymbol{z}_t,\boldsymbol{\theta}_i)\hat{\boldsymbol{\alpha}}_{i,t}$.
The plant is compensated with this unfiltered $\hat{\boldsymbol{h}}_t$.

When the observed residual mixes $\boldsymbol{h}^{\mathrm{reg}}$ with an out-domain residual $\boldsymbol{h}^{\mathrm{off}}$, the raw $\boldsymbol{h}(\boldsymbol{z}_t)$ is not a reliable supervised target.
We therefore low-pass the residual on the short-memory timescale $N_{\mathrm{S}}$,
\begin{equation}
\label{eq:lpf}
\boldsymbol{h}^{\mathrm{filt}}_t
=
\boldsymbol{h}^{\mathrm{filt}}_{t-1}
+
\frac{1}{N_{\mathrm{S}}}
\bigl(\boldsymbol{h}(\boldsymbol{z}_t)-\boldsymbol{h}^{\mathrm{filt}}_{t-1}\bigr),
\end{equation}
initialized by $\boldsymbol{h}^{\mathrm{filt}}_1=\boldsymbol{h}(\boldsymbol{z}_1)$.
The instantaneous loss and packed gradient are
\begin{align}
\label{eq:loss-filt}
l_t(\hat{\sigma}_t,\hat{\boldsymbol{\alpha}}_t)
&\triangleq
\bigl\|\hat{\boldsymbol{h}}_t-\boldsymbol{h}^{\mathrm{filt}}_t\bigr\|^2,\\
g_t
&\triangleq
\bigl[\mathrm{vec}(\nabla_{\hat{\boldsymbol{\alpha}}_t}l_t);\;\nabla_{\hat{\sigma}_t}l_t\bigr]
\in\mathbb{R}^{3M+1}.
\end{align}
The phase lag of~\eqref{eq:lpf} never enters the closed-loop dynamics, because the plant still uses $\hat{\boldsymbol{h}}_t$.

A constant learning step size
$\hat{\sigma}_{t+1}^\prime=\hat{\sigma}_t-\eta\nabla_{\hat{\sigma}_t}l_t$,
$\hat{\boldsymbol{\alpha}}_{t+1}^\prime=\hat{\boldsymbol{\alpha}}_t-\eta\nabla_{\hat{\boldsymbol{\alpha}}_t}l_t$
then treats every $g_t$ equally~\cite{zhou2026efficient}.
That update remains unsafe if $g_t$ is still aligned with an unfinished transient.
Dual Gradient Consensus~(\Cref{alg:DGC}) keeps~\eqref{eq:loss-filt} and the same gradients, replacing $\eta$.

\subsection{Dual Gradient Consensus}\label{subsec:DGC}
The online least-squares step of \Cref{subsec:OLS} forms the filtered residual~\eqref{eq:lpf}, the loss~\eqref{eq:loss-filt}, and the packed gradient $g_t$.
A constant-step-length update of $g_t$, $(\hat{\boldsymbol{\alpha}},\hat{\sigma})\leftarrow(\hat{\boldsymbol{\alpha}},\hat{\sigma})-\eta g_t$, still applies the same $\eta$ to every sample, including transients and abrupt out-domain disturbances for which $g_t$ is not a reliable descent direction.
Dual Gradient Consensus~(\Cref{alg:DGC}) keeps this residual pipeline unchanged and replaces only the step length: it first classifies the adaptation regime, then determines the step length with which $(\hat{\boldsymbol{\alpha}},\hat{\sigma})$ is updated.

DGC stores past gradients in two FIFO buffers.
The long-term gradient memory (LTG)
$\calM^{\mathrm{L}}_{t-1}=\{g_{t-N_{\mathrm{L}}},\ldots,g_{t-1}\}$
(length at most $N_{\mathrm{L}}$)
is the working point of the current domain.
The reference vector is the mean of the stored gradients,
\begin{equation}
\bar{g}^{\mathrm{L}}_{t-1}
=
\frac{1}{|\calM^{\mathrm{L}}_{t-1}|}
\sum_{g\in\calM^{\mathrm{L}}_{t-1}} g,
\end{equation}
\vspace{-2mm}
whenever $\calM^{\mathrm{L}}_{t-1}\neq\emptyset$.
\vspace{1mm}
The short-term gradient memory (STG)
$\calM^{\mathrm{S}}_{t-1}$
(length at most $N_{\mathrm{S}}$)
is the recent trend; its reference $\bar{g}^{\mathrm{S}}_{t-1}$ is likewise the mean of the vectors in $\calM^{\mathrm{S}}_{t-1}$.
The same $N_{\mathrm{S}}$ is the time constant of~\eqref{eq:lpf}.
Let
$\cos(a,b)=a^{\top}b\big/\bigl(\|a\|\|b\|+\varepsilon\bigr)$.
The agreement bit with each memory is
\begin{equation}
\label{eq:signs}
\begin{aligned}
s^{\mathrm{L}}_{t}
&=
\begin{cases}
1, & \cos(g_t,\bar{g}^{\mathrm{L}}_{t-1})>0,\\
0, & \text{otherwise,}
\end{cases}\\[2pt]
s^{\mathrm{S}}_{t}
&=
\begin{cases}
1, & \cos(g_t,\bar{g}^{\mathrm{S}}_{t-1})>0,\\
0, & \text{otherwise.}
\end{cases}
\end{aligned}
\end{equation}
\vspace{-2mm}
computed from the \emph{pre-update} queues.

\vspace{0.5mm}
\begin{definition}[Two-bit classes]\label{def:dgc-regime}
The long-term and short-term memories are taken to be nonempty.
The pair $\bigl(s^{\mathrm{L}}_{t},s^{\mathrm{S}}_{t}\bigr)$ in~\eqref{eq:signs} partitions $g_t$ into four regimes.
If $\bigl(s^{\mathrm{L}}_{t},s^{\mathrm{S}}_{t}\bigr)=(1,1)$, the residual is consistent with the in-domain: DGC keeps LTG and replaces $(\hat{\boldsymbol{\alpha}},\hat{\sigma})$ via $\varphi$.
If $(0,0)$, $g_t$ agrees with neither memory (unfinished shift or noise): DGC keeps LTG and does not replace the parameters.
If $(1,0)$, the working point of the previous domain remains valid while STG has been pulled by a transient: DGC keeps LTG and replaces $(\hat{\boldsymbol{\alpha}},\hat{\sigma})$ via $\varphi$.
If $(0,1)$, a domain shift has completed: DGC hands off $\calM^{\mathrm{L}}\leftarrow\calM^{\mathrm{S}}$, recomputes $\bar{g}^{\mathrm{L}}$, and replaces $(\hat{\boldsymbol{\alpha}},\hat{\sigma})$ via $\varphi$ on the new domain.
\end{definition}

Thus, \Cref{def:dgc-regime} states that $(\hat{\boldsymbol{\alpha}},\hat{\sigma})$ is replaced only when $g_t$ agrees with the (possibly updated) working point, as summarized in \Cref{tab:dgc-regime}.

\begin{table}[t]
\IEEEtabtoppad
\centering
\small
\caption{Two-bit classification of $g_t$ in \Cref{def:dgc-regime}.\\
(agreement is $\cos(\cdot,\cdot)>0$).}
\label{tab:dgc-regime}
\setlength{\tabcolsep}{3pt}
\begin{tabular}{@{}cc l >{\raggedright\arraybackslash}p{2.55cm}@{}}
\toprule
$s^{\mathrm{L}}$ & $s^{\mathrm{S}}$ & Meaning & Action \\
\midrule
$1$ & $1$ & In-domain & Keep LTG; $\varphi$-step \\
$0$ & $0$ & Shift/noise & Keep LTG; no step \\
$1$ & $0$ & Transient & Keep LTG; $\varphi$-step \\
$0$ & $1$ & New domain & LTG$\leftarrow$STG; $\varphi$-step \\
\bottomrule
\end{tabular}
\vspace{-4mm}
\end{table}

After the regime decision, agreement with the (possibly updated) LTG is scored by the confidence map $\varphi$.
Let $\varsigma(x)=(1+e^{-x})^{-1}$. Set $\varphi(c)=0$ for $c\le 0$ and $\varphi(1)=1$; otherwise
\begin{equation}
\label{eq:phi}
\varphi(c)
=
\frac{\varsigma\bigl(k(c-\tau)\bigr)-\varsigma(-k\tau)}
     {\varsigma\bigl(k(1-\tau)\bigr)-\varsigma(-k\tau)}.
\end{equation}
If LTG is empty we set $L_{\mathrm{cons}}=1$ and $L_{\mathrm{loc}}=\mathbf{1}$.
Otherwise
\begin{align}
L_{\mathrm{cons}}
&=
\varphi\bigl(\cos(g_t,\bar{g}^{\mathrm{L}})\bigr),\\
L_{\mathrm{loc},j}
&=
\varphi\!\left(
\frac{g_{t,j}\,\bar{g}^{\mathrm{L}}_{j}}
{|g_{t,j}|\,|\bar{g}^{\mathrm{L}}_{j}|+\varepsilon}
\right)
\end{align}
for $j=1,\ldots,3M$,
i.e.\ $L_{\mathrm{loc}}$ multiplies only the $\boldsymbol{\alpha}$-block.
The resulting rates are
\begin{equation}
\label{eq:rate}
\boldsymbol{\eta}^{(\boldsymbol{\alpha})}_{t}
=
\eta_{\boldsymbol{\alpha}}\,L_{\mathrm{cons}}\,L_{\mathrm{loc}},
\qquad
\eta^{(\sigma)}_{t}
=
\eta_{\sigma}\,L_{\mathrm{cons}}.
\end{equation}
In the $(0,0)$ regime, $\cos(g_t,\bar{g}^{\mathrm{L}})\le 0$ so $L_{\mathrm{cons}}=0$ and the parameters are not replaced;
in the $(0,1)$ regime the handoff makes $\cos(g_t,\bar{g}^{\mathrm{L}})>0$ and learning resumes.

The parameters are then updated by projected gradient descent on the filtered loss~\eqref{eq:loss-filt},
\begin{align}
\hat{\boldsymbol{\alpha}}_{t+1}
&=
\Pi_{\calD_{\boldsymbol{\alpha}}}
\bigl(
\hat{\boldsymbol{\alpha}}_{t}
-
\boldsymbol{\eta}^{(\boldsymbol{\alpha})}_{t}\odot\nabla_{\hat{\boldsymbol{\alpha}}_t}l_t
\bigr),\\
\hat{\sigma}_{t+1}
&=
\Pi_{\calD_{\sigma}}
\bigl(
\hat{\sigma}_{t}
-
\eta^{(\sigma)}_{t}\,\nabla_{\hat{\sigma}_t}l_t
\bigr),
\end{align}
after which $g_t$ is appended to both FIFO memories.
\Cref{tab:dgc-hp} lists the values used in the experiments.

\begin{example}[Wind versus impulse]
A slowly varying wind or payload belongs to $\boldsymbol{h}^{\mathrm{reg}}$: both memories agree with $g_t$, so $\bigl(s^{\mathrm{L}},s^{\mathrm{S}}\bigr)=(1,1)$ and DGC replaces $(\hat{\boldsymbol{\alpha}},\hat{\sigma})$ with a consensus-modulated learning step size.
A short motor-failure spike or collision belongs to $\boldsymbol{h}^{\mathrm{off}}$: $g_t$ agrees with neither memory, so $(0,0)$ and the parameters are not replaced.
If the domain remains valid and only STG is pulled by a transient disturbance or noise, the pair is $(1,0)$: DGC keeps LTG and still takes a $\varphi$-step.
After a completed level change (e.g.\ a new load), STG consolidates first, $(0,1)$ triggers $\mathrm{LTG}\leftarrow\mathrm{STG}$, and resumes on the new domain.
\end{example}

\begin{remark}[Confidence map]
The map \(\varphi\) of~\eqref{eq:phi} converts cosine agreement into a step-size multiplier in \([0,1]\).
The parameters $\tau$ and $k$ only shape the sigmoid; there is no additional hard threshold.
Selecting a single learning step size $\eta$ a priori is difficult when the residual mixes time-varying, stationary, and abrupt out-domain signals.
DGC therefore lets $\varphi$ shrink $\eta$ on weak consensus and restore the base rates $\eta_{\boldsymbol{\alpha}}$, $\eta_{\sigma}$ as soon as the working point is trustworthy.
To avoid shrinking the step on measurement noise, a decrease of $\boldsymbol{\eta}$ is accepted only when the raw prediction error is large:
if $\bigl\|\hat{\boldsymbol{h}}_t-\boldsymbol{h}(\boldsymbol{z}_t)\bigr\|\le\varepsilon_{\min}$ and any coordinate of~\eqref{eq:rate} would fall below the previous applied step size, the previous step size is retained.
The applied step size is the moving average of the last $W$ (held) rates.
\end{remark}

\vspace{-2mm}
\subsection{Algorithm}\label{subsec:algo}
The procedure has an offline stage and an online stage.
Offline, a policy $\pi_{\boldsymbol{\psi}}$ is trained by BPTT as in \Cref{subsec:BPTT}.
Based on the RFF parameterization of \Cref{subsec:RFF}, the dynamics are written
$\boldsymbol{x}_{t+1} = \boldsymbol{f}(\boldsymbol{x}_{t}) + \boldsymbol{g}(\boldsymbol{x}_{t}) \boldsymbol{u}_{t} + \frac{1}{M} \sum_{i=1}^{M} \Phi\left(\boldsymbol{z}_t, \boldsymbol{\theta}_i \right) {\boldsymbol{\alpha}}_{i,k}$.
Each epoch samples $(\sigma_k,\boldsymbol{\alpha}_k)$ from $\calD_\sigma\times\calD_{\boldsymbol{\alpha}}$, rolls out $\pi_{\boldsymbol{\psi}}$ with those kernel parameters in the observation, and updates $\boldsymbol{\psi}$ by backpropagation through the simulator.
The resulting policy is held fixed online.

Online, DAPL~(\Cref{alg:MODA})~\cite{zhou2026efficient} forms the residual gradient and Dual Gradient Consensus~(\Cref{alg:DGC}) replaces its constant step.
\Cref{alg:MODA} is the online phase of DAPL~\cite{zhou2026efficient}: apply $\pi_{\boldsymbol{\psi}}$ with $(\hat{\sigma}_t,\hat{\boldsymbol{\alpha}}_t)$, recover $\boldsymbol{h}(\boldsymbol{z}_t)$, and form $g_t$ from~\eqref{eq:lpf} and~\eqref{eq:loss-filt}.
\Cref{alg:DGC} replaces that step and is the core of the online update.
From the pre-update memories it forms the mean references $\bar{g}^{\mathrm{L}}$ and $\bar{g}^{\mathrm{S}}$, evaluates the agreement pair~\eqref{eq:signs}, and applies the four-regime rule of \Cref{def:dgc-regime}: keep LTG on $(1,1)$, $(0,0)$, and $(1,0)$; hand off $\calM^{\mathrm{L}}\leftarrow\calM^{\mathrm{S}}$ only on $(0,1)$.
The confidence map~\eqref{eq:phi} then converts agreement with the (possibly updated) LTG into the step sizes~\eqref{eq:rate}.
A decrease of $\boldsymbol{\eta}$ is accepted only when the raw residual error exceeds $\varepsilon_{\min}$; the applied learning step size is a moving average of window $W$.
The parameters are replaced by projected gradient descent on~\eqref{eq:loss-filt}, after which $g_t$ is appended to both FIFO memories.
\Cref{tab:dgc-hp} lists the values used in the experiments.

\vspace{-1mm}
\begin{algorithm}[H]
\caption{Domain-Adaptive Policy Learning (DAPL)~\cite{zhou2026efficient} \textit{with residual filtering and Dual Gradient Consensus (Alg.~2)}.}
\begin{algorithmic}[1]
    \REQUIRE Learned policy $\pi_{\boldsymbol{\psi}}$; number of random Fourier features $M$; random Fourier feature parameters $\{\boldsymbol{\theta}\}_{1,\dots,M}$; domain set $\calD_\sigma$ and $\calD_{\boldsymbol{\alpha}}$; base rates $\eta_{\boldsymbol{\alpha}}$, $\eta_{\sigma}$.
    \ENSURE At each time step $t=1,\ldots,T$, control input $u_{t}$.
        \vspace*{-0.6mm}
        \STATE Initialize $\boldsymbol{x}_1$, $\hat{\sigma}_1 \in \calD_\sigma$, $\hat{\boldsymbol\alpha}_{i,1} \in \calD_{\boldsymbol{\alpha}}$, $\calM^{\mathrm{L}}\leftarrow\emptyset$, $\calM^{\mathrm{S}}\leftarrow\emptyset$;
        \vspace*{-3.6mm}
        \STATE Formulate $\boldsymbol\Phi(\cdot, \boldsymbol\theta_i)$, where $i \in \{1, \dots, M\}$;
    \FOR {each time step $t = 1, \dots, T$}
    \STATE Apply control input $u_t = \pi_{\boldsymbol{\psi}}(\boldsymbol{o}_t)$  with $\hat{\sigma}_t$ and $\hat{\boldsymbol\alpha}_{t}$;
        \STATE Observe state $\boldsymbol{x}_{t+1}$, and calculate disturbance via $\boldsymbol{h}(\boldsymbol{z}_t) = \boldsymbol{x}_{t+1} - \boldsymbol{f}(\boldsymbol{x}_{t}) - \boldsymbol{g}(\boldsymbol{x}_{t}) \boldsymbol{u}_{t}$;
        \STATE Predict $\hat{\boldsymbol{h}}_t \leftarrow \frac{1}{M} \sum_{i=1}^{M} \Phi\left(\boldsymbol{z}_t, \boldsymbol{\theta}_i \right) \hat{\boldsymbol{\alpha}}_{i,t}$;
        \IF{$t=1$}
            \STATE $\boldsymbol{h}^{\mathrm{filt}}_t \leftarrow \boldsymbol{h}(\boldsymbol{z}_t)$;
        \ELSE
            \STATE $\boldsymbol{h}^{\mathrm{filt}}_t$ by~\eqref{eq:lpf};
        \ENDIF
        \STATE Set $l_t(\hat{\sigma}_t,\hat{\boldsymbol{\alpha}}_t)\triangleq\|\hat{\boldsymbol{h}}_t-\boldsymbol{h}^{\mathrm{filt}}_t\|^2$;
        \STATE Calculate gradient $\nabla_{\hat{\sigma}_t}$ and $\nabla_{\hat{\boldsymbol{\alpha}}_t}$;
        \STATE Pack $g_t\leftarrow\bigl[\mathrm{vec}(\nabla_{\hat{\boldsymbol{\alpha}}_t}l_t);\;\nabla_{\hat{\sigma}_t}l_t\bigr]$;
        \STATE $(\hat{\boldsymbol{\alpha}}_{t+1},\hat{\sigma}_{t+1},\calM^{\mathrm{L}},\calM^{\mathrm{S}})\leftarrow$ Dual Gradient Consensus~(\Cref{alg:DGC});
        \ENDFOR
\end{algorithmic}\label{alg:MODA}
\end{algorithm}
\vspace{-1mm}

\begin{algorithm}[t]
\IEEEalgtoppad
\caption{Dual Gradient Consensus.}
\begin{algorithmic}[1]
    \REQUIRE $g_t$; $(\hat{\boldsymbol{\alpha}}_t,\hat{\sigma}_t)$; $\calM^{\mathrm{L}}$, $\calM^{\mathrm{S}}$; $\eta_{\boldsymbol{\alpha}}$, $\eta_{\sigma}$; $\tau$, $k$; $N_{\mathrm{L}}$, $N_{\mathrm{S}}$; $\varepsilon_{\min}$; $W$; $\calD_\sigma$, $\calD_{\boldsymbol{\alpha}}$.
    \ENSURE $(\hat{\boldsymbol{\alpha}}_{t+1},\hat{\sigma}_{t+1})$, $\calM^{\mathrm{L}}$, $\calM^{\mathrm{S}}$.
    \medskip
        \STATE $\cos(a,b)\triangleq a^{\top}b/(\|a\|\|b\|+\varepsilon)$,\ $\varsigma(x)\triangleq(1+e^{-x})^{-1}$,\ $\varphi$ as in~\eqref{eq:phi};
        \STATE $c_{j}(g,r)\triangleq g_{j}r_{j}/(|g_{j}|\,|r_{j}|+\varepsilon)$;
        \STATE $\bar{g}^{\mathrm{L}}\leftarrow\frac{1}{|\calM^{\mathrm{L}}|}\sum_{g\in\calM^{\mathrm{L}}}g$ if $\calM^{\mathrm{L}}\neq\emptyset$; likewise $\bar{g}^{\mathrm{S}}$;
        \IF{$\calM^{\mathrm{L}}=\emptyset$}
            \STATE $L_{\mathrm{cons}}\leftarrow 1$,\ $L_{\mathrm{loc}}\leftarrow\mathbf{1}$;
        \ELSE
            \STATE $c^{\mathrm{L}}\leftarrow\cos(g_t,\bar{g}^{\mathrm{L}})$;\ $c^{\mathrm{S}}\leftarrow\cos(g_t,\bar{g}^{\mathrm{S}})$ if $\calM^{\mathrm{S}}\neq\emptyset$, else $0$;
            \IF{$c^{\mathrm{L}}\le 0$ \textbf{and} $c^{\mathrm{S}}>0$}
                \STATE \textit{new domain:} $\calM^{\mathrm{L}}\leftarrow\calM^{\mathrm{S}}$, recompute $\bar{g}^{\mathrm{L}}$, $c^{\mathrm{L}}\leftarrow\cos(g_t,\bar{g}^{\mathrm{L}})$;
            \ELSE
                \STATE \textit{otherwise} keep $\calM^{\mathrm{L}}$ (in-domain, shift/noise, or out-domain);
            \ENDIF
            \STATE $L_{\mathrm{cons}}\leftarrow\varphi(c^{\mathrm{L}})$;\ $L_{\mathrm{loc},j}\leftarrow\varphi\bigl(c_{j}(g_t,\bar{g}^{\mathrm{L}})\bigr)$;
        \ENDIF
        \STATE $\boldsymbol{\eta}^{(\boldsymbol{\alpha}),\mathrm{raw}}\leftarrow\eta_{\boldsymbol{\alpha}}\,L_{\mathrm{cons}}\,L_{\mathrm{loc}}$,\ $\eta^{(\sigma),\mathrm{raw}}\leftarrow\eta_{\sigma}\,L_{\mathrm{cons}}$;
        \STATE $\boldsymbol{\eta}^{\mathrm{h}}\leftarrow\boldsymbol{\eta}_{t-1}$ if $\bigl\|\hat{\boldsymbol{h}}_t-\boldsymbol{h}(\boldsymbol{z}_t)\bigr\|\le\varepsilon_{\min}$ and $\boldsymbol{\eta}^{\mathrm{raw}}\lneq\boldsymbol{\eta}_{t-1}$; else $\boldsymbol{\eta}^{\mathrm{raw}}$;
        \STATE $\boldsymbol{\eta}_t\leftarrow\frac{1}{W_t}\sum_{i=0}^{W_t-1}\boldsymbol{\eta}^{\mathrm{h}}_{t-i}$,\ $W_t=\min(W,t)$;
        \STATE $\hat{\boldsymbol{\alpha}}_{t+1}=\Pi_{\calD_{\boldsymbol{\alpha}}}\bigl(\hat{\boldsymbol{\alpha}}_{t}-\boldsymbol{\eta}^{(\boldsymbol{\alpha})}_{t}\odot\nabla_{\hat{\boldsymbol{\alpha}}_t}l_t\bigr)$,\ $\hat{\sigma}_{t+1}=\Pi_{\calD_{\sigma}}\bigl(\hat{\sigma}_{t}-\eta^{(\sigma)}_{t}\nabla_{\hat{\sigma}_t}l_t\bigr)$;
        \STATE $\calM^{\mathrm{L}}\leftarrow\mathrm{FIFO}_{N_{\mathrm{L}}}(\calM^{\mathrm{L}},g_t)$,\ $\calM^{\mathrm{S}}\leftarrow\mathrm{FIFO}_{N_{\mathrm{S}}}(\calM^{\mathrm{S}},g_t)$.
\end{algorithmic}\label{alg:DGC}
\end{algorithm}

\begin{table}[t]
\IEEEtabtoppad
\centering
\small
\caption{Hyper-parameters of Dual Gradient Consensus.}
\label{tab:dgc-hp}
\setlength{\tabcolsep}{4pt}
\begin{tabular}{@{}l p{3.6cm} c@{}}
\toprule
Symbol & Meaning & Value \\
\midrule
$N_{\mathrm{L}}$ & LTG length & $20$ \\
$N_{\mathrm{S}}$ & STG length and LPF~\eqref{eq:lpf} & $5$ \\
$\tau$ & Center of $\varphi$ & $0.5$ \\
$k$ & Slope of $\varphi$ & $10$ \\
$\varepsilon_{\min}$ & Hold threshold on raw $\|\hat{\boldsymbol{h}}-\boldsymbol{h}\|$ & $0.2$ \\
$W$ & Moving-average window of $\eta$ & $10$ \\
$\eta_{\boldsymbol{\alpha}},\eta_{\sigma}$ & Base rates in~\eqref{eq:rate} & $0.1$ \\
\bottomrule
\end{tabular}
\vspace{-6mm}
\end{table}

%% file: sec/4-Exp.tex
\section{Experiments}
\label{sec:exp}
In this section we report the experimental evaluation.
We first describe the training protocol and the algorithm parameters in~\Cref{subsec:train}, then the benchmark controllers in~\Cref{subsec:benchmark}, and the simulations in~\Cref{subsec:sim}.

\subsection{Policy Training and Algorithm Parameters}\label{subsec:train}
\begin{figure*}[t]
\centering
\includegraphics[width=\textwidth]{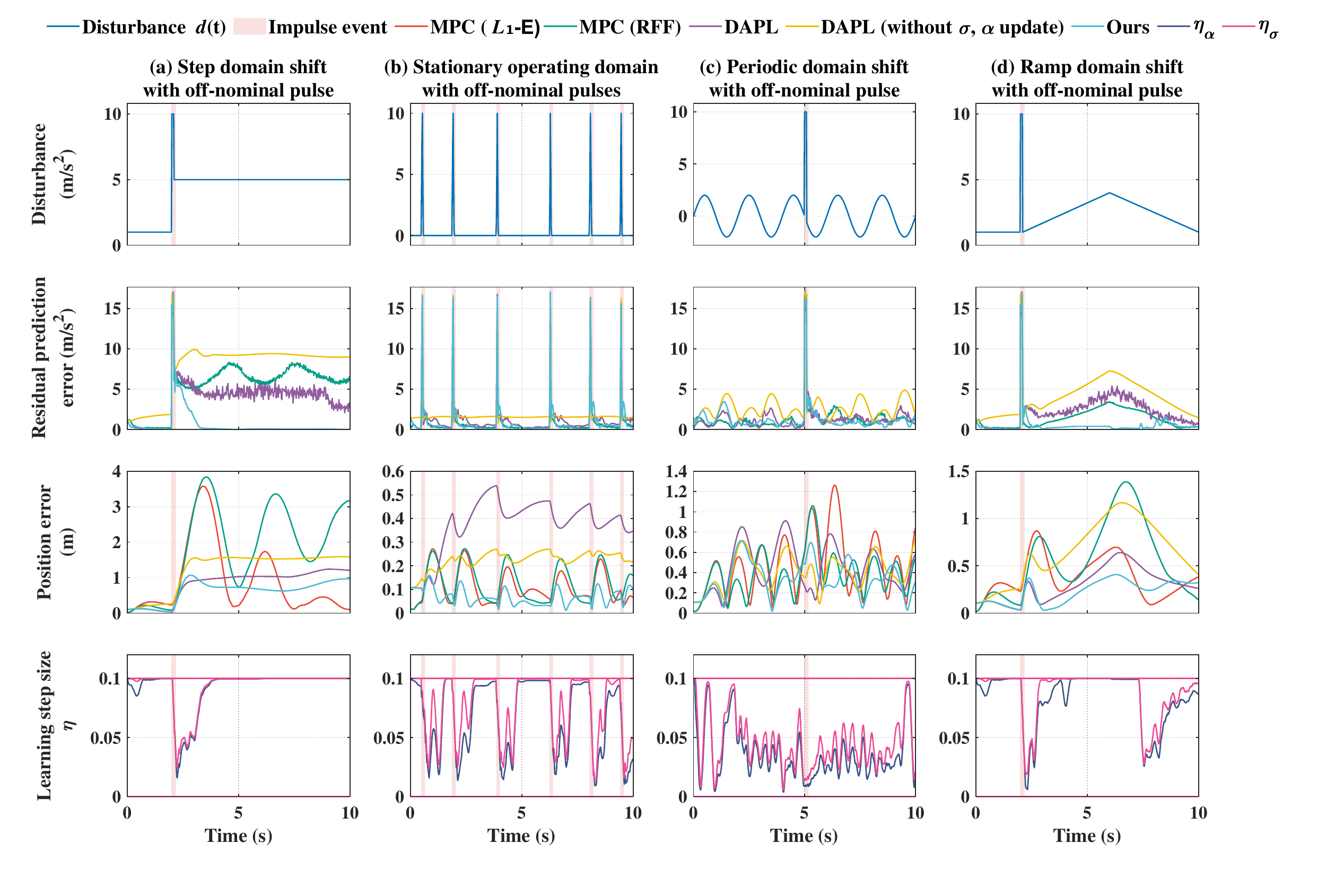}
\vspace{-8mm}
\caption{Representative rollouts for the four cases in~\Cref{subsec:sim}.
Columns:
(a)~step domain shift with out-domain pulse (Case~1);
(b)~stationary domain with out-domain pulses (Case~2);
(c)~periodic domain shift with out-domain pulse (Case~3);
(d)~ramp domain shift with out-domain pulse (Case~4).
Rows, top to bottom: applied disturbances (one coordinate, $\mathrm{m/s}^{2}$), residual prediction error ($\mathrm{m/s}^{2}$), position tracking error ($\mathrm{m}$), and Dual Gradient Consensus rates $\eta_{\boldsymbol{\alpha}}$ and $\eta_{\sigma}$.
Shaded bands mark impulse intervals.
Controllers: MPC~(L1-E), MPC~(RFF), DAPL, DAPL~(w/o $\sigma$, $\boldsymbol{\alpha}$ update), and DGC (ours).}
\label{fig:sim-result}
\vspace{-4mm}
\end{figure*}
The policy is a three-layer multilayer perceptron with hidden width $512$.
Its observation $\boldsymbol{o}_t\in\mathbb{R}^{187}$ stacks the body-frame velocity $\boldsymbol{R}^{\top}\boldsymbol{v}_t$, the quaternion $\boldsymbol{q}_t$, the previous thrust $T_{t-1}$, the previous angular-velocity command $\boldsymbol{\omega}_{t-1}$, ten relative future reference points $\Delta\boldsymbol{x}$ over a $1.0\,\mathrm{s}$ horizon, and the kernel parameters $\boldsymbol{\alpha}_t$ and $\sigma_t$.
The network outputs mass-normalized thrust $T_t/m$ and body-rate $\boldsymbol{\omega}_t$.

Offline training uses BPTT~(\Cref{subsec:BPTT}) on a lemniscate tracking task.
We train in JAX with $500$ parallel environments for $300$ epochs.
Each episode lasts at most $250$ simulation steps at $\Delta t=0.02\,\mathrm{s}$.
The policy step size starts at $0.001$ and follows a cosine decay.
At the beginning of every episode we sample $\boldsymbol{\alpha}\sim\mathcal{N}(\boldsymbol{0},\tfrac{1}{2}\boldsymbol{I})$ and $\sigma$ uniformly from $[0.001,\,1]$.
We use a Gaussian kernel with $M=25$ random features, hence $\boldsymbol{\alpha}\in\mathbb{R}^{75}$.
The reward is the negative of the position, quaternion, and velocity tracking errors together with the deviation of the command from hover, weighted $2.5$, $0.5$, $0.1$, and $0.01$.
Training finishes in $50\,\mathrm{s}$ on an RTX~$4090$.

Online, $\boldsymbol{\alpha}$ and $\sigma$ are replaced by Dual Gradient Consensus~(\Cref{alg:DGC}) with the hyper-parameters in~\Cref{tab:dgc-hp}.
We initialize $\boldsymbol{\alpha}$ at zero and $\sigma=0.5$.
The RFF input $\boldsymbol{z}$ comprises $\boldsymbol{R}^{\top}\boldsymbol{v}_t$, $\boldsymbol{q}_t$, the normalized thrust $T_{t-1}/mg$, and $\boldsymbol{\omega}_{t-1}$.
The base rates are $\eta_{\boldsymbol{\alpha}}=\eta_{\sigma}=0.1$.

\subsection{Benchmark Algorithms}\label{subsec:benchmark}
We compare five algorithms, in the same top-to-bottom order as~\Cref{table:sim-pos}: MPC~(L1-E)~\cite{huang2023datt}, MPC~(RFF)~\cite{zhou2025simultaneous}, Domain-Adaptive Policy Learning~(DAPL)~\cite{zhou2026efficient}, DAPL~(w/o $\sigma$, $\boldsymbol{\alpha}$ update), and ours.

MPC~(L1-E) uses model predictive control on the quadrotor dynamics~\eqref{eq:quadrotor} and augments the nominal model with an $L_1$ state predictor~\cite{wu2023mathcal}.
MPC~(RFF)~\cite{zhou2025simultaneous} parametrizes the residual by random Fourier features, injects it into sampling-based MPC, and updates $\boldsymbol{\alpha}$ and $\sigma$ online.
DAPL~\cite{zhou2026efficient} is the constant-rate domain-adaptive policy from previous work: it forms the filtered least-squares loss of~\Cref{subsec:OLS} and applies a fixed step to both $\hat{\boldsymbol{\alpha}}$ and $\hat{\sigma}$.
DAPL~(w/o $\sigma$, $\boldsymbol{\alpha}$ update) uses the same predictor but never replaces $(\hat{\boldsymbol{\alpha}},\hat{\sigma})$ after initialization.

Ours uses the identical policy, features, and loss as DAPL; the only change is that~\Cref{alg:DGC} classifies the residual regime and, through the confidence map $\varphi$, selects the step length of the $(\hat{\boldsymbol{\alpha}},\hat{\sigma})$ update.

\subsection{Numerical Simulations}\label{subsec:sim}
We evaluate $10\,\mathrm{s}$ rollouts under four residuals that mix an in-domain residual $\boldsymbol{h}^{\mathrm{reg}}$ with an out-domain residual $\boldsymbol{h}^{\mathrm{off}}$, as in~\Cref{prob:control}.
The applied $\boldsymbol{h}$ is a world-frame acceleration residual ($\mathrm{m/s}^{2}$), identical on all three axes; \Cref{fig:sim-result} plots one coordinate.
\begin{enumerate}[label=(\alph*)]
    \item Step domain shift with out-domain pulse (Case~1): $\boldsymbol{h}^{\mathrm{reg}}$ is $1\,\mathrm{m/s}^{2}$ until $t=2\,\mathrm{s}$ and $5\,\mathrm{m/s}^{2}$ thereafter; $\boldsymbol{h}^{\mathrm{off}}$ is a $0.1\,\mathrm{s}$ pulse of $10\,\mathrm{m/s}^{2}$ on $[2.0,\,2.1)\,\mathrm{s}$.
    \item Stationary domain with out-domain pulses (Case~2): $\boldsymbol{h}^{\mathrm{reg}}=\boldsymbol{0}$; $\boldsymbol{h}^{\mathrm{off}}$ comprises six pulses of peak $10\,\mathrm{m/s}^{2}$ and width $0.06$--$0.12\,\mathrm{s}$, separated by at least $1.2\,\mathrm{s}$.
    \item Periodic domain shift with out-domain pulse (Case~3): $\boldsymbol{h}^{\mathrm{reg}}(t)=2\sin(\pi t)\,\mathrm{m/s}^{2}$; a $0.1\,\mathrm{s}$ pulse of $10\,\mathrm{m/s}^{2}$ replaces the sinusoid on $[5.0,\,5.1)\,\mathrm{s}$.
    \item Ramp domain shift with out-domain pulse (Case~4): $\boldsymbol{h}^{\mathrm{reg}}$ is $1\,\mathrm{m/s}^{2}$ until $t=2\,\mathrm{s}$, then a triangle $1\to 4\to 1\,\mathrm{m/s}^{2}$ on $[2.1,\,6.0]$ and $[6.0,\,10]\,\mathrm{s}$; $\boldsymbol{h}^{\mathrm{off}}$ is a $0.1\,\mathrm{s}$ pulse of $10\,\mathrm{m/s}^{2}$ on $[2.0,\,2.1)\,\mathrm{s}$.
\end{enumerate}
The performance metric is position RMSE, reported as mean $\pm$ standard deviation over ten rollouts in~\Cref{table:sim-pos}.
\Cref{fig:sim-result} shows a representative $10\,\mathrm{s}$ rollout for each case.
Columns~(a)--(d) follow Cases~1--4.
From top to bottom the rows are the applied disturbances (one world-frame coordinate), the residual prediction error, the position tracking error, and the Dual Gradient Consensus rates $\eta_{\boldsymbol{\alpha}}$ and $\eta_{\sigma}$, with shaded intervals marking $\boldsymbol{h}^{\mathrm{off}}$.

Among the five controllers, Dual Gradient Consensus has the lowest mean RMSE in every column of~\Cref{table:sim-pos}.
The comparison is qualitative as well as quantitative: the third row of~\Cref{fig:sim-result} is consistent with the table, and the fourth row shows when the proposed update reduces the step.
Where the residual prediction error spikes on a shaded transient, the Dual Gradient Consensus rates fall: those samples are not written into $(\hat{\boldsymbol{\alpha}},\hat{\sigma})$.

\begingroup
\setlength{\parskip}{0pt}
On Case~1~(\Cref{fig:sim-result}(a)) the prediction error spikes on the pulse at $t=2\,\mathrm{s}$ and the step length drops, so $(\hat{\boldsymbol{\alpha}},\hat{\sigma})$ are not overwritten; after the hold they follow the new level and the prediction error returns toward zero.
MPC~(RFF) and MPC~(L1-E) oscillate after the level change ($5.660\pm 1.508\,\mathrm{m}$ and $1.921\pm 0.524\,\mathrm{m}$).
DAPL~(w/o $\sigma$, $\boldsymbol{\alpha}$ update) cannot track the new residual; constant-step DAPL tracks it only in part ($0.929\pm 0.525\,\mathrm{m}$ vs.\ $0.478\pm 0.006\,\mathrm{m}$).

On Case~2~(\Cref{fig:sim-result}(b)) each spike raises the prediction error and Dual Gradient Consensus lowers $\boldsymbol{\eta}$ on the shaded intervals, remaining at $0.006\pm 0.001\,\mathrm{m}$.
DAPL writes the spikes into $(\hat{\boldsymbol{\alpha}},\hat{\sigma})$ and reaches $0.162\pm 0.044\,\mathrm{m}$.
DAPL~(w/o $\sigma$, $\boldsymbol{\alpha}$ update) keeps a large prediction error; both MPC variants stay accurate ($0.017\pm 0.002\,\mathrm{m}$ and $0.022\pm 0.002\,\mathrm{m}$).

On Case~3~(\Cref{fig:sim-result}(c)) the rates vary with the sinusoid; the prediction error peaks on the mid-horizon pulse and $\boldsymbol{\eta}$ dips, so that transient is filtered.
Dual Gradient Consensus has a lower peak position error than the MPC baselines and lower RMSE than DAPL ($0.145\pm 0.004\,\mathrm{m}$ vs.\ $0.255\pm 0.056\,\mathrm{m}$).

On Case~4~(\Cref{fig:sim-result}(d)) the rates drop on the pulse at $t=2\,\mathrm{s}$ while the prediction error is large, then recover on the triangle-shaped disturbances ($0.068\pm 0.010\,\mathrm{m}$).
DAPL~(w/o $\sigma$, $\boldsymbol{\alpha}$ update) cannot follow it ($0.531\pm 0.001\,\mathrm{m}$); MPC~(RFF) degrades on some rollouts ($0.565\pm 0.496\,\mathrm{m}$).
\par
\endgroup

%% file: sec/5-Conclusion.tex
\begin{table}[t]
\IEEEtabtoppad
\centering
\small
\caption{Position RMSE (m) for the four cases in~\Cref{subsec:sim}, mean $\pm$ standard deviation over ten rollouts.
Best mean in each column in \textbf{bold}.}
\label{table:sim-pos}
\setlength{\tabcolsep}{3pt}
\newcommand{\msd}[2]{\begin{tabular}{@{}c@{}}#1\\[-1pt]{\scriptsize$\pm #2$}\end{tabular}}
\begin{tabular}{@{}lcccc@{}}
\toprule
Case & 1 & 2 & 3 & 4 \\
\midrule
MPC~(L1-E) & \msd{$1.921$}{0.524} & \msd{$0.017$}{0.002} & \msd{$0.312$}{0.037} & \msd{$0.190$}{0.030} \\
MPC~(RFF) & \msd{$5.660$}{1.508} & \msd{$0.022$}{0.002} & \msd{$0.207$}{0.036} & \msd{$0.565$}{0.496} \\
DAPL & \msd{$0.929$}{0.525} & \msd{$0.162$}{0.044} & \msd{$0.255$}{0.056} & \msd{$0.133$}{0.077} \\
\begin{tabular}[t]{@{}l@{}}DAPL~(w/o $\sigma$,\\$\boldsymbol{\alpha}$ update)\end{tabular}
 & \msd{$1.812$}{0.001} & \msd{$0.052$}{0.002} & \msd{$0.191$}{0.001} & \msd{$0.531$}{0.001} \\
Ours & \msd{$\mathbf{0.478}$}{0.006} & \msd{$\mathbf{0.006}$}{0.001} & \msd{$\mathbf{0.145}$}{0.004} & \msd{$\mathbf{0.068}$}{0.010} \\
\bottomrule
\end{tabular}
\vspace{-4mm}
\end{table}

\vspace{-2mm}
\section{Limitations}
\label{sec:limitations}
\vspace{-0.5pt}
The residual prediction error $\hat{\boldsymbol{h}}_t-\boldsymbol{h}(\boldsymbol{z}_t)$ is used only as a scalar hold on the Dual Gradient Consensus step size.
It is not treated as a signal from which high-frequency disturbance content could be extracted or rejected.
DGC therefore does not apply a high-frequency filter to that error: impulsive or other broadband transients are handled only through consensus of the two gradient memories.
The residual is already low-pass filtered in~\eqref{eq:lpf}. A duplex that selects the better of Dual Gradient Consensus and constant-step DAPL could distinguish high-frequency residual changes without sacrificing online learning.
The evaluation in this paper is numerical only, and we do not provide a closed-loop stability argument for the gated step length.

\section{Conclusion}
\label{sec:conclusion}
We presented Dual Gradient Consensus for online adaptive learning on a quadrotor.
The method retains the random Fourier feature model, the offline BPTT policy, and the least-squares loss of Domain-Adaptive Policy Learning, and revises only the step length of the residual-gradient update.
Two gradient memories classify each packed gradient and, through the confidence map $\varphi$, set that step length: domain shifts remain trackable, while short out-domain impulses no longer dictate the update of $(\hat{\boldsymbol{\alpha}},\hat{\sigma})$.
In numerical simulation, four residuals that mix an in-domain component with an impulse were compared against MPC~(L1-E), MPC~(RFF), DAPL~\cite{zhou2026efficient}, and DAPL without $\sigma$ and $\boldsymbol{\alpha}$ updates.
Dual Gradient Consensus attained the lowest mean position RMSE in each case; the difference was largest when a constant-rate update absorbed repeated spikes.

%% file: sec/6-Appendix.tex
\useRomanappendicesfalse
\appendices

\section{Benchmark Algorithm Implementation}\label{sec:implementation}

\subsection{MPC~(L1-E)}
This baseline pairs the sampling-based MPC of~\cite{huang2023datt} with the $L_1$ predictor of~\cite{wu2023mathcal}.
Each sampled trajectory carries position, velocity, quaternion, and the extra lag states of the discrete plant; the decision variables are collective thrust and body rate.
We plan $H=40$ steps ($0.8\,\mathrm{s}$ at $\Delta t=0.02\,\mathrm{s}$) and, at every control instant, score $1024$ perturbed sequences with softmax temperature $0.1$.
Command noise has standard deviation $[0.25,\,1,\,1,\,1]$ on weight-normalized thrust and the three body rates; roll and pitch are limited to $[-6,6]\,\mathrm{rad/s}$ and yaw to $[-4,4]\,\mathrm{rad/s}$.
The forward model includes first-order thrust and rate lag with coefficient $0.4$.
A quadratic stage cost penalizes position, attitude, velocity, and deviation from hover with weights $2.5$, $0.5$, $0.1$, and $0.1$.
The $L_1$ loop uses $A=-0.01$, a $0.02\,\mathrm{s}$ update, and low-pass coefficient $0.99$; the estimate is fed to the MPC each step.

\subsection{MPC~(RFF)}
The sampler, horizon, costs, and limits are those of MPC~(L1-E); only the residual channel changes, from the $L_1$ predictor to the RFF learner of~\cite{zhou2025simultaneous}.
Features are built from an $11$-vector---quaternion, body-frame velocity, $T/mg$, and the body-rate command---using $M=25$ random Fourier terms.
Frequencies are drawn i.i.d.\ from a Gaussian of width $\sigma=0.5$ and phases uniformly from $[0,2\pi]$.
The coefficient $\boldsymbol{\alpha}\in\mathbb{R}^{25\times 3}$ starts at zero; $\boldsymbol{\alpha}$ and $\sigma$ are both updated at every control step with rate $0.1$, with $\sigma$ initialized at $0.5$.
The resulting $\hat{\boldsymbol{h}}$ is added to the nominal dynamics.

\subsection{\texorpdfstring{DAPL and DAPL~(w/o $\sigma$, $\boldsymbol{\alpha}$ update)}{DAPL and DAPL (w/o sigma, alpha update)}}
DAPL is the constant-step online phase of~\cite{zhou2026efficient}, stated as~\Cref{alg:MODA}.
The BPTT policy of~\Cref{subsec:train} is run with the current $(\hat{\sigma}_t,\hat{\boldsymbol{\alpha}}_t)$ stacked in the observation.
At each step we recover $\boldsymbol{h}(\boldsymbol{z}_t)$ from~\eqref{eq:residual-ood}, form the filtered label~\eqref{eq:lpf} and the loss~\eqref{eq:loss-filt}, and take a projected step of size $\eta=0.1$ on $(\hat{\boldsymbol{\alpha}},\hat{\sigma})$.
The RFF coordinates $\boldsymbol{z}$ and the count $M=25$ match~\Cref{subsec:train}; the online run starts from $\hat{\boldsymbol{\alpha}}=\boldsymbol{0}$ and $\hat{\sigma}=0.5$.
The frozen baseline keeps that policy, predictor, and observation, but holds $(\hat{\boldsymbol{\alpha}},\hat{\sigma})$ at the initial values for the entire rollout.

\vspace{-2.5mm}
\subsection{Ours}
DGC uses DAPL's policy, features, and loss.
Alg.~\ref{alg:DGC} (Table~\ref{tab:dgc-hp}) sets $N_{\mathrm{L}}=20$, $N_{\mathrm{S}}=5$, $\tau=0.5$, $k=10$, $\varepsilon_{\min}=0.2$, with window $W=10$ and step lengths $\eta_{\boldsymbol{\alpha}}=\eta_{\sigma}=0.1$.